%% file: main.tex
\documentclass[10pt,twocolumn,letterpaper]{article}

\usepackage[pagenumbers]{cvpr} 

\usepackage{colortbl}
\usepackage{graphicx}
\usepackage{amsmath}
\usepackage{amssymb}
\usepackage{booktabs}
\usepackage{times}
\usepackage{epsfig}
\usepackage{xcolor}
\usepackage{comment}
\usepackage{multirow}
\usepackage{boldline}
\usepackage{algpseudocode}
\usepackage{algorithm}
\usepackage[accsupp]{axessibility} 

\algnewcommand\algorithmicgiven{\textbf{Given:}}
\algnewcommand\Given{\item[\algorithmicgiven]}

\definecolor{cvprblue}{rgb}{0.21,0.49,0.74}
\definecolor{Gray}{gray}{0.9}
\usepackage[pagebackref,breaklinks,colorlinks,citecolor=cvprblue]{hyperref}

\def\paperID{16764} 
\def\confName{CVPR}
\def\confYear{2024}

\begin{document}


\title{GreedyViG: Dynamic Axial Graph Construction for Efficient Vision GNNs}


\author{Mustafa Munir, William Avery, Md Mostafijur Rahman, and Radu Marculescu\\
The University of Texas at Austin\\
Austin, Texas, USA\\
{\tt\small \{mmunir, williamaavery, mostafijur.rahman, radum\}@utexas.edu}
}

\maketitle

\def\thefootnote{$^1$}\footnotetext{Code: \url{https://github.com/SLDGroup/GreedyViG}.}


\begin{abstract}
Vision graph neural networks (ViG) offer a new avenue for exploration in computer vision. A major bottleneck in ViGs is the inefficient k-nearest neighbor (KNN) operation used for graph construction. To solve this issue, we propose a new method for designing ViGs, Dynamic Axial Graph Construction (DAGC), which is more efficient than KNN as it limits the number of considered graph connections made within an image. Additionally, we propose a novel CNN-GNN architecture, GreedyViG, which uses DAGC. Extensive experiments show that GreedyViG beats existing ViG, CNN, and ViT architectures in terms of accuracy, GMACs, and parameters on image classification, object detection, instance segmentation, and semantic segmentation tasks. Our smallest model, GreedyViG-S, achieves 81.1\% top-1 accuracy on ImageNet-1K, 2.9\% higher than Vision GNN and 2.2\% higher than Vision HyperGraph Neural Network (ViHGNN), with less GMACs and a similar number of parameters. Our largest model, GreedyViG-B obtains 83.9\% top-1 accuracy, 0.2\% higher than Vision GNN, with a 66.6\% decrease in parameters and a 69\% decrease in GMACs. GreedyViG-B also obtains the same accuracy as ViHGNN with a 67.3\% decrease in parameters and a 71.3\% decrease in GMACs. Our work shows that hybrid CNN-GNN architectures not only provide a new avenue for designing efficient models, but that they can also exceed the performance of current state-of-the-art models$^1$.

\end{abstract}

\section{Introduction}

\begin{figure}[h]
\centering
\includegraphics[width=\columnwidth]{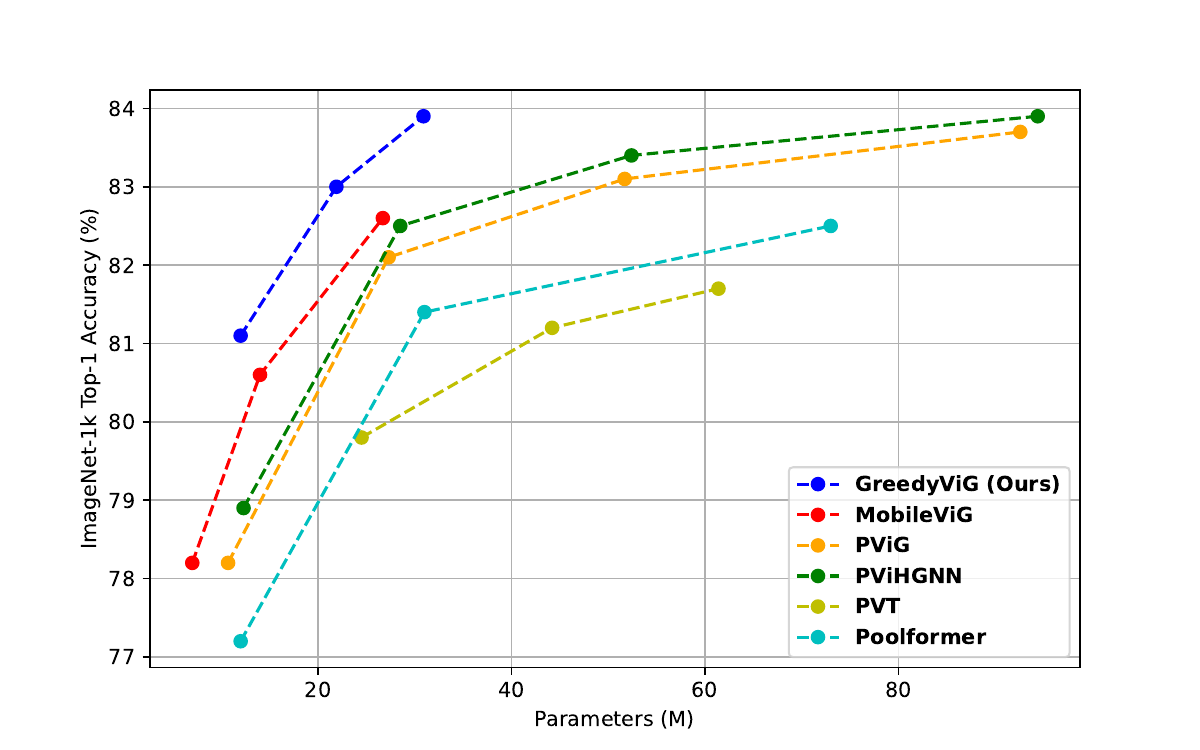}
\caption{\textbf{Comparison of model size and performance (top-1 accuracy on ImageNet-1K)}. GreedyViG achieves the highest performance compared to other state-of-the-art models.} 
\label{fig:pareto}
\end{figure}

Rapid growth in deep learning has lead to numerous successes across a diverse set of computer vision tasks including image classification \cite{imagenet1k}, object detection \cite{coco}, instance segmentation \cite{coco}, and semantic segmentation \cite{ADE20K}. Key drivers behind this growth include convolutional neural networks (CNNs) \cite{lecun1998gradient, Alexnet2012, Resnet, liu2022convnet}, vision transformers (ViTs) \cite{ViT, carion2020end}, and multi-layer perceptron (MLP)-based vision models \cite{mlpmixer, resmlp}. In CNNs and MLPs input images are represented as a grid of pixels, however in ViTs images are represented as a sequence of patches. By splitting an input image into a sequence of patch embeddings, the image is transformed into an input usable by the transformer modules often used in natural language processing \cite{vaswani2017attention, liu2021swin}. Unlike CNNs and MLPs, which have a local receptive field, ViTs have global receptive fields allowing them to learn from distant interactions within images \cite{ViT}.

 The recently proposed Vision GNN (ViG) \cite{Vision_GNN} represents images in a more versatile manner through a graph structure rather than as a sequence of patches as in ViTs \cite{ViT}. ViG constructs the graph through dividing an image into patches and then connecting the patches (i.e., nodes) through the K-nearest neighbors (KNN) algorithm \cite{Vision_GNN}. Vision HyperGraph Neural Network (ViHGNN) \cite{ViHGNN} improves upon the original ViG by using the hypergraph structure to remove the constraint of exclusively connecting pairs of nodes. Like ViTs, ViG-based models can process global object interactions, but are also computationally expensive. To deal with the computationally expensive nature of ViG-based models, MobileViG \cite{MobileViG} used a structured graph that does not change across input images and removes the need for KNN-based graph construction.

While the success of ViG \cite{Vision_GNN}, ViHGNN \cite{ViHGNN}, and MobileViG \cite{MobileViG} show the potential of treating an image as a graph for computer vision tasks, they also show some limitations. In general ViG-based models are computationally expensive, due to the expensive nature of graph construction. MobileViG alleviates this issue through static graph construction, but at the cost of a graph that does not change across input images, thus limiting the benefit of using a graph-based model. The limitations of current ViG-based models are as follows:

\begin{enumerate}
    \item \textbf{Computational Cost of Graph Construction:} A fundamental issue facing ViG-based models is the cost of KNN-based graph construction. KNN-based graph construction requires comparing every single node within the ViG-based model to determine the $K$ nearest nodes. This cost makes KNN-based ViG models inefficient.
    \item \textbf{Inability of a Static Graph to Change Across Inputs:} The computational cost of KNN-based ViG models lead to static graph construction. The fundamental issue with static graph construction is it removes the benefit of using a ViG-based model as the graph constructed no longer changes across input images. 
\end{enumerate}

 In this work, we propose Dynamic Axial Graph Construction (DAGC) to address the current limitations of ViG-based models. We also introduce GreedyViG, an efficient ViG-based architecture using a hybrid CNN-GNN approach. DAGC is more computationally efficient compared to KNN-based graph construction while maintaining a dynamic set of connections that changes across input images. In Figure \ref{fig:pareto}, we show that our proposed GreedyViG architecture outperforms competing state-of-the-art (SOTA) models across all model sizes in terms of parameters. We summarize our contributions as follows:

\begin{enumerate}
    \item We propose a new method for designing efficient vision GNNs, Dynamic Axial Graph Construction (DAGC). DAGC is more efficient compared to KNN-based ViGs as DAGC limits the graph connections made within an image to only the most significant ones. Our method is lightweight compared to KNN-based ViGs and more representative than SOTA static graph construction based ViGs.
    \item We propose a novel efficient CNN-GNN architecture, GreedyViG, which uses DAGC, conditional positional encoding (CPE) \cite{CPE}, and max-relative graph convolution \cite{maxrel}. We use convolutional layers and grapher layers in all four stages of the proposed architecture to perform local and global processing for each resolution.
    \item We conduct comprehensive experiments to underscore the efficacy of the GreedyViG architecture, which beats existing ViG architectures, efficient CNN architectures, and efficient ViT architectures in terms of accuracy and/or parameters and GMACs (number of MACs in billions) on four representative vision tasks: ImageNet image classification \cite{imagenet1k}, COCO object detection \cite{coco}, COCO instance segmentation \cite{coco}, and ADE20K semantic segmentation \cite{ADE20K}. Specifically our GreedyViG-B model achieves a top-1 accuracy of 83.9\% on the ImageNet classification task, 46.3\% Average Precision ($AP$) on the COCO object detection task, and 47.4\% mean Intersection over Union ($mIoU$) on the ADE20K semantic segmentation task.
\end{enumerate}

The paper is organized as follows. Section \ref{Sec:Rel_Work} covers related work in the ViG and efficient computer vision architecture space. Section \ref{Sec:Methodology} describes the design methodology behind DAGC and the GreedyViG architecture. Section \ref{Sec:Results} describes experimental setup and results for ImageNet-1k image classification, COCO object detection, COCO instance segmentation, and ADE20K semantic segmentation. Section \ref{Sec:Ablations} covers ablation studies on how different design decisions impact performance on ImageNet-1k. Lastly, Section \ref{Sec:Conclusion} summarizes our main contributions.

\section{Related Work}
\label{Sec:Rel_Work}

The mainstream network architecture for computer vision has historically been convolutional neural networks (CNN) \cite{Alexnet2012, lecun1998gradient, Resnet, VGGNet, Densenet}. In the efficient computer vision space, CNN-based architectures \cite{MobileNet, MobileNetv2, tan2019efficientnet, tan2021efficientnetv2} have been even more dominant due to the computationally expensive nature of ViTs \cite{ViT}. Many works have attempted to address the computational costs associated with self-attention layers \cite{pan2022hilo, wang2020linformer} and recently hybrid architectures that combine CNNs and ViTs to effectively capture local and global information have been proposed \cite{FastViT, EfficientFormer, EfficientFormerv2, MobileViT, MobileViTv2}. 

Traditionally graph neural networks (GNNs) have operated on biological, social, or citation networks \cite{kipf2016semi, hamilton2017inductive, zhou2020graph, wu2020graph}. GNNs have also been used for tasks in computer vision such as, point cloud classification and segmentation \cite{landrieu2018large, wang2019dynamic}, as well as human action recognition \cite{yan2018spatial}. But, with the introduction of Vision GNN \cite{Vision_GNN}, the adoption of GNNs as a general purpose vision backbone has grown with works like \cite{ViHGNN, PVG, MobileViG}. MobileViG \cite{MobileViG} introduces a hybrid CNN-GNN architecture to design an efficient computer vision backbone to compete with CNN, ViT, and hybrid architectures. MobileViG accomplishes this through introducing a static graph construction method called Sparse Vision Graph Attention (SVGA) to avoid the computationally expensive nature of ViGs. Despite the efficiencies of MobileViG \cite{MobileViG}, it does not take full advantage of the global processing possible with GNNs since it only uses graph convolution at the lowest resolution stage of its design. MobileViG \cite{MobileViG} also loses representation ability because all images construct the same graph in their proposed static method, decreasing the benefits of using a GNN-based architecture. Thus, to address these limitations, we introduce a new CNN-GNN architecture, GreedyViG, that takes advantage of graph convolution at higher resolution stages and constructs a graph that changes across input images.

\begin{figure}[h]
\centering
\includegraphics[width=\columnwidth]{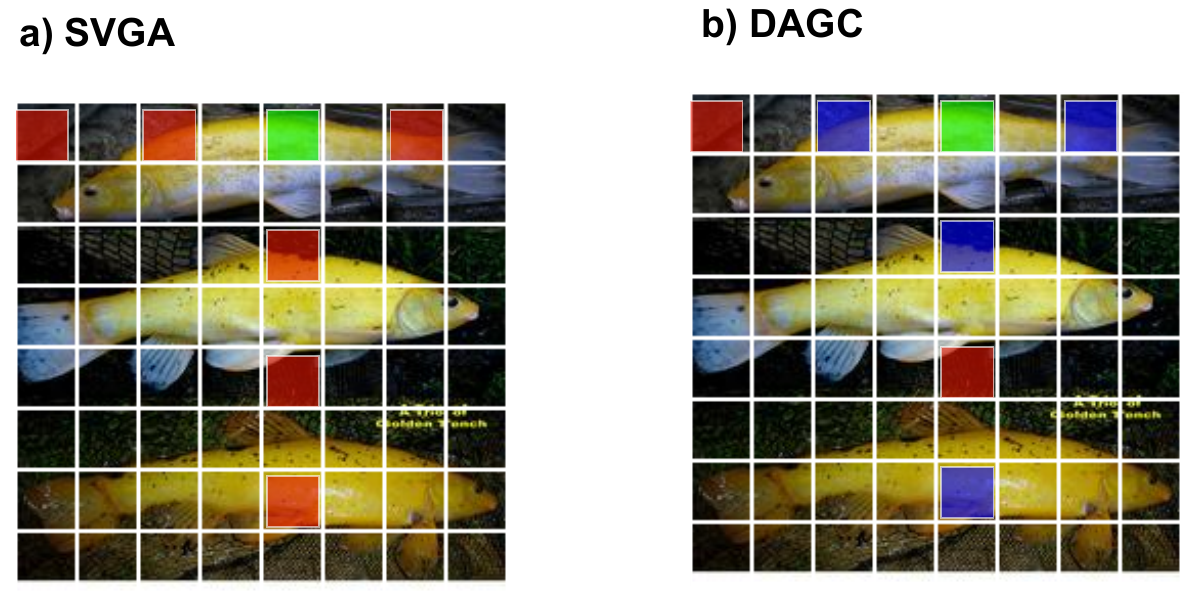}
\caption{\textbf{DAGC and SVGA graph construction.} a) SVGA graph construction for the green patch of an 8$\times$8 image. All red patches will be connected to the green patch regardless of similarity. b) DAGC for the green patch of an 8$\times$8 image. DAGC dynamically constructs a graph along the axes, through applying a mask (the blue patches) to only connect similar patches in terms of Euclidean distance. The red patches will not be connected to the green patch as they are not a part of the mask.} 
\label{fig:KNN_ViG}
\end{figure}

\section{Methodology}
\label{Sec:Methodology}

In this section, we describe the DAGC algorithm and provide details on the GreedyViG architecture design. More precisely, Section 3.1 describes the DAGC algorithm. Section 3.2 explains how we adapt the Grapher module from ViG \cite{Vision_GNN} to create the DAGC block. Section 3.3 describes how we combine the DAGC blocks along with inverted residual blocks \cite{MobileNetv2} to create the GreedyViG architecture.

\subsection{Dynamic Axial Graph Construction}

We propose Dynamic Axial Graph Construction (DAGC) as an efficient alternative dynamic graph construction method to the computationally expensive KNN graph construction method from Vision GNN \cite{Vision_GNN}. DAGC builds upon SVGA \cite{MobileViG}, but instead of statically constructing a graph, DAGC constructs a graph that changes across input images. DAGC retains the efficiencies of SVGA through the removal of the KNN computation and input reshaping. It also introduces an efficient graph construction method based on the mean ($\mu$) and standard deviation ($\sigma$) of the Euclidean distance between patches in the input image.

In ViG, the KNN computation is required for every input image, since the nearest neighbors of each patch cannot be known ahead of time. This results in a graph with connections throughout the image. Due to the unstructured nature of KNN, ViG \cite{Vision_GNN} contains two reshaping operations. The first to reshape the input image from a 4D tensor to a 3D tensor for graph convolution and the second to reshape the input from 3D back to 4D for the convolutional layers. SVGA \cite{MobileViG} eliminates these two reshaping operations and KNN computation through using a static graph where each patch is connected to every $K^{th}$ patch in its row and column as seen in Figure \ref{fig:KNN_ViG}a.


DAGC leverages the axial construction of SVGA to retain its efficiencies, while dynamically constructing a more representative graph. To do this, DAGC first obtains an estimate of the $\mu$ and $\sigma$ of the Euclidean distance between nodes through using a subset of nodes. The subset of nodes is obtained by splitting the image into quadrants and comparing the quadrants diagonal to one another as shown in Figure \ref{fig:quadrants} below. Then, the $\mu$ and $\sigma$ can be calculated with those Euclidean distance values. This allows the estimated $\mu$ and $\sigma$ to be computed between two images (the original and the one with its quadrants flipped across the diagonal). This is to decrease the number of comparisons for getting the $\mu$ and $\sigma$ values. The reason we avoid calculating the true $\mu$ and $\sigma$ is that computing them directly would require calculating the Euclidean distance between each individual node and all other nodes in the image. We then consider connections across the row and column as in SVGA to decrease computation, as MobileViG \cite{MobileViG} demonstrated that not every patch needs to be considered. If the Euclidean distance between two nodes is less than the difference of the estimated $\mu$ and $\sigma$, then we connect the two nodes.

\begin{figure}[h]
\centering
\includegraphics[width=\columnwidth]{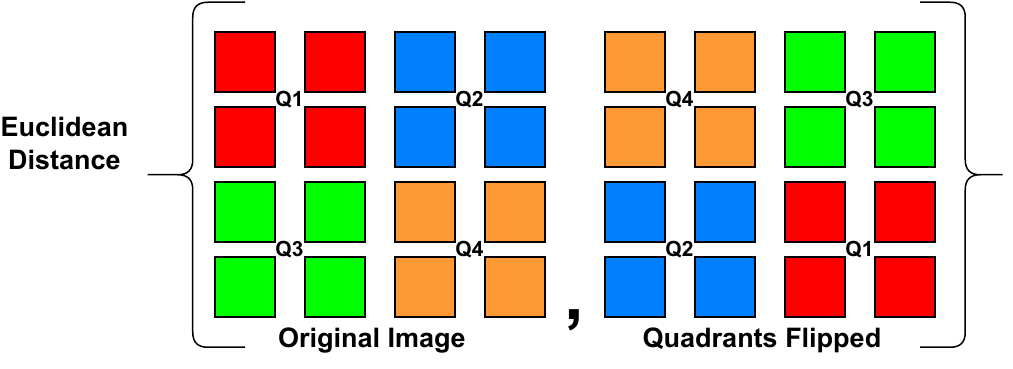}
\caption{\textbf{Euclidean distance calculation} between the original image and the image with its quadrants flipped along the diagonal.} 
\label{fig:quadrants}
\end{figure}


\begin{figure*}[t]
\centering
\includegraphics[scale=0.725]{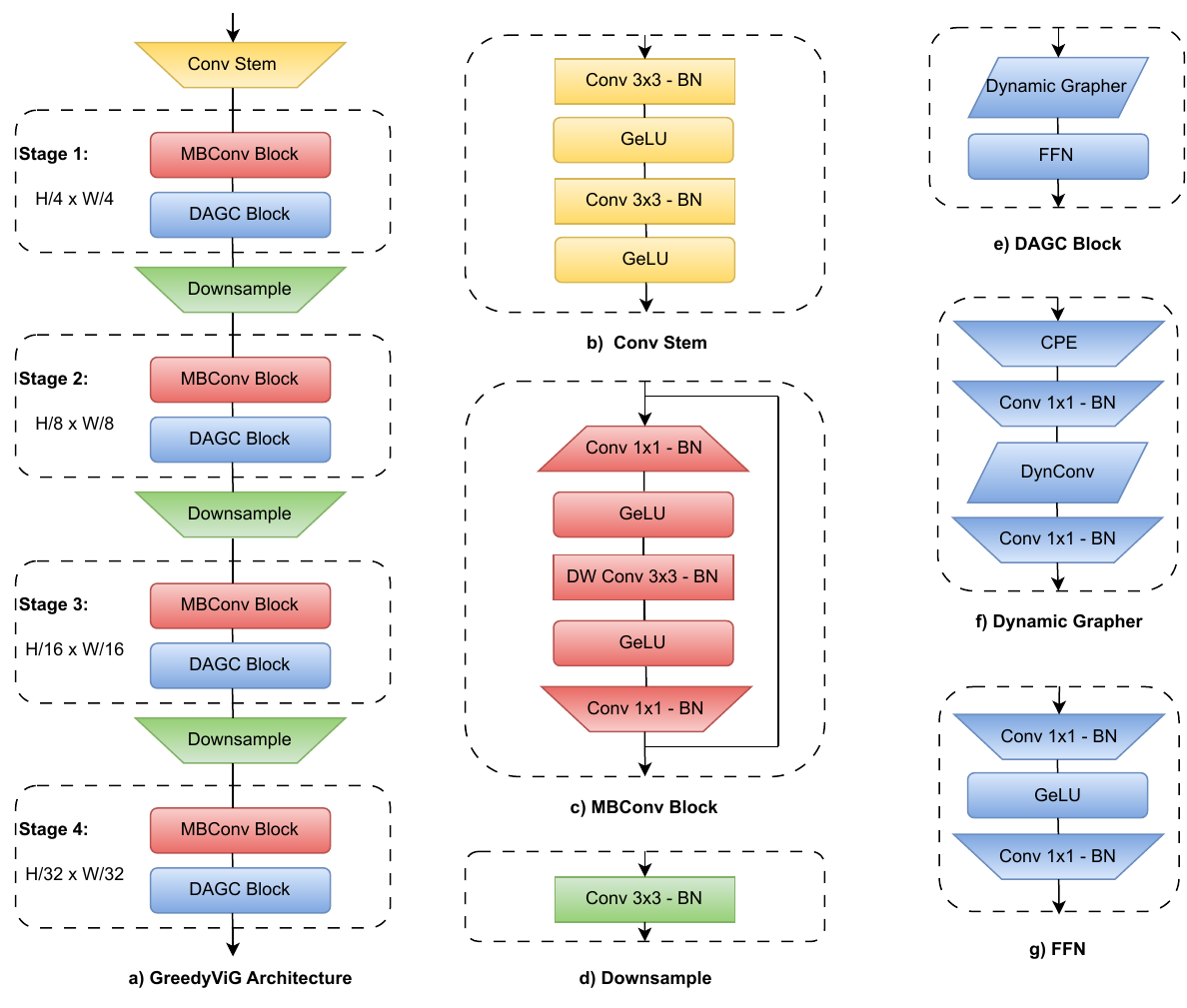}
\caption{\textbf{GreedyViG architecture.} (a) Network architecture showing the stages and blocks. (b) The Conv Stem. (c) MBConv Block. (d) Downsample. (e) DAGC Block. (f) Dynamic Grapher. (g) FFN.}
\label{fig:GreedyViG_Architecture}
\end{figure*}

DAGC also enables a variable amount of connections across different images, unlike KNN's fixed $K$ number of connections for all images. This is because in different images, different nodes will have a Euclidean distance between them be less than the difference of the estimated $\mu$ and $\sigma$. The intuition behind the use of $\mu$ and $\sigma$ is that node pairs that are within one $\sigma$ of $\mu$ are close to one another and should share information. These values are then used to make the connections generated by DAGC as shown in Figure \ref{fig:KNN_ViG}b. In Figure \ref{fig:KNN_ViG} we can see that SVGA connects the fish to parts of the image that are not fish, while DAGC only connects the fish to other parts of the image that are fish.

Now that we have the estimated $\mu$ and $\sigma$ within the image, we $roll$ the input image $X$, $mK$ to the $right$ or $down$ while $mK$ is less than $H$ (height) and $W$ (width) of the image as seen in Algorithm \ref{alg:cap}. The $roll$ operation is used to compare the patches that are $N$ hops away. In Figure \ref{fig:KNN_ViG}b, node (5,1) in \textit{x, y} coordinates is compared to nodes (1,1), (3,1), (7,1), (5,3), (5,5), and (5,7) through "rolling" to the next node. After rolling, we compute the Euclidean distance between the input $X$ and the rolled version ($X_{rolled}$) to determine whether to connect the two points. If the distance is less than $\mu - \sigma$, then the mask is assigned a value of 1, else it is assigned a value of 0. This mask is then multiplied with $X_{rolled}-X$, to mask out max-relative scores between patches not considered connections. This value is denoted as $X_{down}$ and $X_{right}$ in Algorithm \ref{alg:cap}. Next, the $max$ operation is taken and the result is stored in $X_{final}$. Lastly, after the rolling, masking, and max-relative operations a final $Conv2d$ is performed.

Through this methodology, DAGC provides a more representative graph construction compared to SVGA \cite{MobileViG}, as dissimilar patches (i.e., nodes) are not connected. DAGC is also less computationally expensive compared to KNN due to less comparisons being needed for constructing the graph (KNN must compute the nearest neighbors for every patch). DAGC also does not require the reshaping needed for performing graph convolution in KNN-based methods \cite{Vision_GNN}. Thus, DAGC provides representation flexibility like KNN and decreased computational complexity like SVGA. 


\begin{algorithm}
\caption{DAGC}\label{alg:cap}
\begin{algorithmic}
\Given $K$, the distance between connections; $H,W$, the image resolution; $X$, the input image; $X_{quadrants}$, the quadrants of the input flipped across the diagonals; $m$, the distance of each roll.

\State $m \gets 0$
\State $norm \gets norm(X, X_{quadrants})$ \Comment{matrix norm of tensors}
\State $\mu \gets mean(norm)$
\State $\sigma \gets std(norm)$

\While{$mK < H$}
    \State $X_{rolled} \gets roll_{down}(X, mK))$
    \State $dist \gets norm(X, X_{rolled})$     \Comment{get distance value}
    \If{$dist < \mu- \sigma$}   \Comment{generate mask}
        \State $mask \gets 1$
    \Else
        \State $mask \gets 0$
    \EndIf
    \State $X_{down} \gets mask * (X_{rolled}-X)$ \Comment{get features}
    \State $X_{final} \gets max(X_{down}, X_{final})$ \Comment{keep max}
    \State $m \gets m+1$            
\EndWhile
\State $m \gets 0$
\While{$mK < W$}
    \State $X_{rolled} \gets roll_{right}(X, mK)$
    \State $dist \gets norm(X, X_{rolled})$  
    \If{$dist < \mu - \sigma$} 
        \State $mask \gets 1$
    \Else
        \State $mask \gets 0$
    \EndIf
    \State $X_{right} \gets mask * (X_{rolled}-X)$  
    \State $X_{final} \gets max(X_{right}, X_{final})$
    \State $m \gets m+1$
\EndWhile
\State \Return $Conv2d(Concat(X,X_{final}))$
\end{algorithmic}
\end{algorithm}

\subsection{DAGC Block}

The DAGC block consists of a Dynamic Grapher module followed by a feed-forward network (FFN). The Dynamic Grapher module differs from the Grapher module used in \cite{MobileViG} through the use of an updated max-relative graph convolution step called DynConv using Algorithm \ref{alg:cap} and conditional positional encoding (CPE) \cite{CPE}. DynConv dynamically creates a graph that changes across input images, unlike the graph construction used in the graph convolution of SVGA \cite{MobileViG}. Given an input feature $X\in\mathbb{R}^{N \times N}$, the updated Dynamic Grapher is expressed as:

\begin{equation}
\tag{1}
Y=\sigma(DynConv((X + CPE(X))W_{in}))W_{out}+X
\end{equation}

where $Y\in\mathbb{R}^{N \times N}$, $W_{in}$ and $W_{out}$ are fully connected layer weights, $CPE$ is a depthwise convolution, and $\sigma$ is a GeLU activation. The updated Dynamic Grapher module is visually depicted in Figure \ref{fig:GreedyViG_Architecture}f.

Following the updated Dynamic Grapher, we use the feed-forward network (FFN) module as used in Vision GNN \cite{Vision_GNN} and MobileViG \cite{MobileViG}, which can be seen in Figure \ref{fig:GreedyViG_Architecture}g. The FFN module is a two layer MLP expressed as:
\begin{equation}
\tag{2}
Z=\sigma(XW_{1})W_{2}+Y
\end{equation}
where $Z\in\mathbb{R}^{N \times N}$, $W_{1}$ and $W_{2}$ are fully connected layer weights, and $\sigma$ is once again GeLU. We call this combination of the Dynamic Grapher module and FFN the DAGC block, as shown in Figure \ref{fig:GreedyViG_Architecture}e.

CPE is introduced into the DAGC block to provide the position of the node within the image as this is important to performance \cite{islam2019much, CPE}. The CPE used in DAGC follows the method of \cite{CPE}, i.e., a depthwise convolution computes the encodings, and then the encodings are added to the input. The addition of CPE adds spatial information into the message passing step of DynConv improving performance.

\subsection{GreedyViG Architecture}

The GreedyViG architecture shown in Figure \ref{fig:GreedyViG_Architecture}a is composed of a convolutional stem, and four stages of inverted residual blocks (MBConv) and DAGC blocks each followed by a downsample reducing the resolution to get to the next stage. The stem consists of 3$\times$3 convolutions with the stride equal to 2, each followed by batch normalization (BN) and the GeLU activation function as seen in Figure \ref{fig:GreedyViG_Architecture}b. The MBConv block is used for local processing at each stage, before the DAGC block performs global processing at each stage. Each MBConv block consists of pointwise convolutions, BN, GeLU, a depth-wise 3$\times$3 convolution, and a residual connection as seen in Figure \ref{fig:GreedyViG_Architecture}c. The DAGC block is used at each resolution to better learn global object interactions. Between each stage there is a downsample, which consists of a 3$\times$3 convolution with a stride equal to 2 followed by BN as shown in Figure \ref{fig:GreedyViG_Architecture}d to half the input resolution and expand the channel dimension. Each stage in the GreedyViG architecture is composed of multiple MBConv and DAGC blocks, where the number of repetitions and channel width is changed depending on model size. Within the DAGC blocks used in all GreedyViG model sizes, the distance between connections of nodes before masking is set to $K$ = 8, 4, 2, 1 for stages 1 to 4, respectively. This allows the graph constructed to still be dense in lower resolution stages, as too sparse of a graph can negatively impact accuracy as seen Table \ref{tab:ablation2} in our ablation studies. After the final DAGC block there is a classification head consisting of Average Pooling and an FFN.

\begin{table*}[ht]
\fontsize{10.5}{11}\selectfont
\def\arraystretch{1.3}
\caption{\textbf{Classification results on ImageNet-1k} for GreedyViG and other state-of-the-art models. Different training seeds result in about 0.1\% variation in accuracy for GreedyViG over three runs. Bold entries indicate results obtained for GreedyViG proposed in this paper.}
\centering
\begin{tabular}[t]{|c|c|c|c|c|c|c|c|c|}
\hline
{\textbf{Model}} & {\textbf{Type}} & {\textbf{Parameters (M)}} & {\textbf{GMACs}} & {\textbf{Epochs}} & {\textbf{Top-1 Accuracy (\%)}}        \\ \hline

ResNet18 \cite{Resnet} & CNN & 11.7 & 1.82 & 300 & 69.7 \\ \hline
ResNet50 \cite{Resnet}                       & CNN          & 25.6      & 4.1       & 300 & 80.4      \\ \hline
ConvNext-T \cite{liu2022convnet}                     & CNN          & 28.6      & 7.4        & 300 & 82.7      \\ \hlineB{5}
EfficientFormer-L1 \cite{EfficientFormer}             & CNN-ViT        & 12.3      & 1.3      & 300 & 79.2     \\ \hline
EfficientFormer-L3 \cite{EfficientFormer}              & CNN-ViT        & 31.3      & 3.9    & 300 & 82.4      \\ \hline
EfficientFormer-L7 \cite{EfficientFormer}              & CNN-ViT        & 82.1      & 10.2   & 300 & 83.3 
    \\ \hline
LeViT-192 \cite{graham2021levit}            & CNN-ViT        & 10.9      & 0.7        & 1000 & 80.0     \\ \hline
LeViT-384 \cite{graham2021levit}                      & CNN-ViT        & 39.1      & 2.4        & 1000 & 82.6     \\ \hline
EfficientFormerV2-S2 \cite{EfficientFormerv2}       & CNN-ViT    & 12.6     & 1.3      & 300    & 81.6 \\ \hline
EfficientFormerV2-L \cite{EfficientFormerv2}             & CNN-ViT        & 26.1      & 2.6      & 300 & 83.3          \\ \hlineB{5}
PVT-Small \cite{wang2021pyramid}             & ViT         & 24.5          & 3.8       & 300 & 79.8 \\ \hline
PVT-Large \cite{wang2021pyramid}             & ViT         & 61.4          & 9.8       & 300 & 81.7 \\ \hline
DeiT-S \cite{Deit}                         & ViT     & 22.5      & 4.5       & 300 & 81.2      \\ \hline
Swin-T \cite{liu2021swin}                      & ViT     & 29.0      & 4.5        & 300 & 81.4      \\ \hline
PoolFormer-s12 \cite{MetaFormer}            & Pool          & 12.0      & 2.0        & 300 & 77.2     \\ \hline
PoolFormer-s24 \cite{MetaFormer}                 & Pool          & 21.0      & 3.6        & 300 & 80.3      \\ \hline 
PoolFormer-s36 \cite{MetaFormer}                 & Pool          & 31.0      & 5.2         & 300 & 81.4       \\ \hlineB{5}
PViHGNN-Ti  \cite{ViHGNN}       & GNN & 12.3     & 2.3 & 300 & 78.9       \\ \hline
PViHGNN-S    \cite{ViHGNN}     & GNN & 28.5     & 6.3 & 300 & 82.5      \\ \hline
PViHGNN-B    \cite{ViHGNN}     & GNN & 94.4 & 18.1     & 300 & 83.9     \\ \hline
PViG-Ti \cite{Vision_GNN}          & GNN & 10.7     & 1.7 & 300 & 78.2       \\ \hline
PViG-S \cite{Vision_GNN}          & GNN & 27.3     & 4.6 & 300 & 82.1      \\ \hline
PViG-B \cite{Vision_GNN}          & GNN & 92.6 & 16.8     & 300 & 83.7     \\ \hlineB{5}
{MobileViG-S} \cite{MobileViG}    & {CNN-GNN}        & {7.2}       & {1.0}     & {300} & {78.2}       \\ \hline
{MobileViG-M} \cite{MobileViG}     & {CNN-GNN}        & {14.0}         & {1.5}   & {300} & {80.6}       \\ \hline
{MobileViG-B}  \cite{MobileViG}    & {CNN-GNN}        & {26.7}         & {2.8}     & {300} & {82.6}       \\ \hlineB{5}
\rowcolor {Gray}
\textbf{GreedyViG-S (Ours)}     & \textbf{CNN-GNN}        & \textbf{12.0}       & \textbf{1.6}     & \textbf{300} & \textbf{81.1}       \\ \hline
\rowcolor {Gray}
\textbf{GreedyViG-M (Ours)}     & \textbf{CNN-GNN}        & \textbf{21.9}         & \textbf{3.2}   & \textbf{300} & \textbf{82.9}       \\ \hline
\rowcolor {Gray}
\textbf{GreedyViG-B (Ours)}     & \textbf{CNN-GNN}        & \textbf{30.9}         & \textbf{5.2}     & \textbf{300} & \textbf{83.9}       \\ \hlineB{5}
\end{tabular}
\label{Classification_Results}
\end{table*}

\begin{table*}[ht]
\small 
\def\arraystretch{1.2}
\caption{\textbf{Object detection, instance segmentation, and semantic segmentation results} of GreedyViG and other backbones on MS COCO 2017 and ADE20K. (-) denotes unrevealed or unsupported models. Bold entries indicate results obtained using GreedyViG and DAGC proposed in this paper.}
\centering
\begin{tabular}[t]{|c|c|c|c|c|c|c|c|c|c|c|}
\hline
{\textbf{Backbone}} & {\textbf{Parameters (M)}} & \textbf{$AP^{box}$} & \textbf{$AP^{box}_{50}$} & \textbf{$AP^{box}_{75}$} & \textbf{$AP^{mask}$} & \textbf{$AP^{mask}_{50}$} & \textbf{$AP^{mask}_{75}$} & \textbf{$mIoU$} \\ \hline
                    
ResNet18 \cite{Resnet}        & 11.7  & 34.0 & 54.0 & 36.7 & 31.2 & 51.0 & 32.7 & 32.9    \\ \hline
EfficientFormer-L1 \cite{EfficientFormer}   & 12.3 & 37.9 & 60.3 & 41.0 & 35.4 & 57.3 & 37.3 & 38.9       \\ \hline
EfficientFormerV2-S2 \cite{EfficientFormerv2}     & 12.6 & 43.4 & 65.4 & 47.5 & 39.5 & 62.4 & 42.2 & 42.4     \\ \hline
PoolFormer-S12 \cite{MetaFormer}   & 12.0  & 37.3 & 59.0 & 40.1 & 34.6 & 55.8 & 36.9   & 37.2        \\ \hline
FastViT-SA12 \cite{FastViT}  & 10.9 & 38.9 & 60.5 & 42.2 & 35.9 & 57.6 & 38.1 & 38.0       \\ \hline
{MobileViG-M} \cite{MobileViG}    & {14.0} & {41.3} & {62.8} & {45.1} & {38.1} & {60.1} & {40.8} & -   \\ \hline
\rowcolor {Gray}
\textbf{GreedyViG-S (Ours)}    & \textbf{12.0} & \textbf{43.2} & \textbf{65.2} & \textbf{47.3} & \textbf{39.8} & \textbf{62.2} & \textbf{43.2} & \textbf{43.2}    \\ \hlineB{5}

ResNet50 \cite{Resnet}            & 25.5 & 38.0 & 58.6 & 41.4 & 34.4 & 55.1 & 36.7 & 36.7    \\ \hline
EfficientFormer-L3 \cite{EfficientFormer}    & 31.3 & 41.4 & 63.9 & 44.7 & 38.1 & 61.0 & 40.4 & 43.5     \\ \hline
EfficientFormer-L7 \cite{EfficientFormer}    & 82.1 &  42.6 & 65.1 & 46.1 & 39.0 & 62.2 & 41.7 & 45.1       \\ \hline
EfficientFormerV2-L \cite{EfficientFormerv2}      & 26.1 & 44.7 & 66.3 & 48.8 & 40.4 & 63.5 & 43.2 & 45.2   \\ \hline
PoolFormer-S24 \cite{MetaFormer}          & 21.0 & 40.1 & 62.2 & 43.4 & 37.0 & 59.1 & 39.6 & 40.3    \\ \hline
FastViT-SA36 \cite{FastViT}                   & 30.4 & 43.8 & 65.1 & 47.9 & 39.4 & 62.0 & 42.3 & 42.9 \\ \hline
Pyramid ViG-S \cite{Vision_GNN}       & 27.3 & 42.6 & 65.2 & 46.0 & 39.4 & 62.4 & 41.6 & -   \\ \hline
Pyramid ViHGNN-S \cite{ViHGNN}           & 28.5 & 43.1 & 66.0 & 46.5 & 39.6 & 63.0 & 42.3 & - \\ \hline
PVT-Small \cite{wang2021pyramid}         & 24.5 & 40.4 & 62.9 & 43.8 & 37.8 & 60.1 & 40.3  & 39.8   \\ \hline
{MobileViG-B} \cite{MobileViG}   & {26.7} & {42.0} & {64.3} & {46.0} & {38.9} & {61.4} & {41.6} & -   \\ \hline
\rowcolor {Gray}
\textbf{GreedyViG-B (Ours)}    & \textbf{30.9} & \textbf{46.3} & \textbf{68.4} & \textbf{51.3} & \textbf{42.1} & \textbf{65.5} & \textbf{45.4}  & \textbf{47.4}    \\ \hlineB{5}
\end{tabular}
\label{Object_Detection_Segmentation_Results}
\end{table*}

\section{Experimental Results}
\label{Sec:Results}

We compare GreedyViG to ViG \cite{Vision_GNN}, ViHGNN \cite{ViHGNN}, MobileViG \cite{MobileViG}, and other efficient vision architectures to show that for each model size, GreedyViG has a superior performance on the tasks of image classification, object detection, instance segmentation, and semantic segmentation for similar or less parameters and GMACs.

\subsection{Image Classification}

We implement the model using PyTorch 1.12.1 \cite{paszke2019pytorch} and Timm library \cite{timm}. We use 16 NVIDIA A100 GPUs to train our models, with an effective batch size of 2048. The models are trained from scratch for 300 epochs on ImageNet-1K \cite{imagenet1k} with AdamW optimizer \cite{AdamW}. Learning rate is set to 2e$^{-3}$ with cosine annealing schedule. We use a standard image resolution, 224 × 224, for both training and testing. Similar to DeiT \cite{Deit}, we perform knowledge distillation using RegNetY-16GF \cite{RegNetY} with 82.9\% top-1 accuracy. 

As seen in Table \ref{Classification_Results}, for a similar number of parameters and GMACs, GreedyViG outperforms Pyramid ViG (PViG) \cite{Vision_GNN}, Pyramid ViHGNN (PViHGNN) \cite{ViHGNN}, and MobileViG \cite{MobileViG} significantly. For example, our smallest model, GreedyViG-S, achieves 81.1\% top-1 accuracy on ImageNet-1K with 12.0 M parameters and 1.6 GMACs, which is 2.9\% higher top-1 accuracy compared to PViG-Ti \cite{Vision_GNN} and 2.2\% higher than PViHGNN-Ti \cite{ViHGNN} with less GMACs and a similar number of parameters. Our largest model, GreedyViG-B obtains 83.9\% top-1 accuracy with only 30.9 M parameters and 5.2 GMACs, which is a 0.2\% higher top-1 accuracy compared to PViG-B \cite{Vision_GNN} with a 66.6\% decrease in parameters (61.7 M fewer parameters) and a 69\% decrease in GMACs (11.6 fewer GMACs) and the same top-1 accuracy as PViHGNN-B \cite{ViHGNN} with a 67.3\% decrease in parameters (63.5 M fewer parameters) and a 71.3\% decrease in GMACs (12.9 fewer GMACs).

When compared to other efficient architectures in Table \ref{Classification_Results}, GreedyViG beats SOTA models in accuracy for a similar number of parameters and GMACs. GreedyViG-S beats PoolFormer-s12 \cite{MetaFormer} with 3.9\% higher top-1 accuracy while having the same number of parameters and 0.4 fewer GMACs. GreedyViG-M achieves 82.9\% top-1 accuracy beating ConvNext-T \cite{liu2022convnet} with 0.2\% higher top-1 accuracy while having 6.7 M fewer parameters and 4.2 fewer GMACs. Additionally, GreedyViG-B achieves 83.9\% top-1 accuracy beating the EfficientFormer \cite{EfficientFormer, EfficientFormerv2} family of models for a similar number of parameters.

\subsection{Object Detection and Instance Segmentation}

We show that GreedyViG generalizes well to downstream tasks by using it as a backbone in the Mask-RCNN framework \cite{mask_r_cnn} for object detection and instance segmentation tasks on the MS COCO 2017 dataset \cite{coco}. The dataset contains training and validations sets of 118K and 5K images, respectively. We implement the backbone using PyTorch 1.12.1 \cite{paszke2019pytorch} and Timm library \cite{timm}. The model is initialized with ImageNet-1k pretrained weights from 300 epochs of training. We use the AdamW \cite{Adam, AdamW} optimizer with an initial learning rate of 2e$^{-4}$ and train the model for 12 epochs with a standard resolution (1333 $\times$ 800) following the process of prior work \cite{li2022next, MobileViG, EfficientFormer, EfficientFormerv2}.

As seen in Table \ref{Object_Detection_Segmentation_Results}, with similar model size GreedyViG outperforms PoolFormer \cite{MetaFormer}, EfficientFormer \cite{EfficientFormer}, EfficientFormerV2 \cite{EfficientFormerv2}, MobileViG \cite{MobileViG}, and PVT \cite{wang2021pyramid} in terms of either parameters or improved average precision ($AP$) on object detection and instance segmentation. The GreedyViG-S model gets 43.2 $AP^{box}$ and 39.8 $AP^{mask}$ on the object detection and instance segmentation tasks outperforming PoolFormer-s12 \cite{MetaFormer} by 5.9 $AP^{box}$ and 5.2 $AP^{mask}$. Our GreedyViG-B model achieves 46.3 $AP^{box}$ and 42.1 $AP^{mask}$ outperforming MobileViG-B \cite{MobileViG} by 4.3 $AP^{box}$ and 3.2 $AP^{mask}$ and FastViT-SA36 \cite{FastViT} by 2.5 $AP^{box}$ and 2.7 $AP^{mask}$. The strong performance of GreedyViG on object detection and instance segmentation shows the capability of DAGC and GreedyViG to generalize well to different tasks in computer vision.

\subsection{Semantic Segmentation}

\begin{figure}[h]
\centering
\includegraphics[width=\columnwidth]{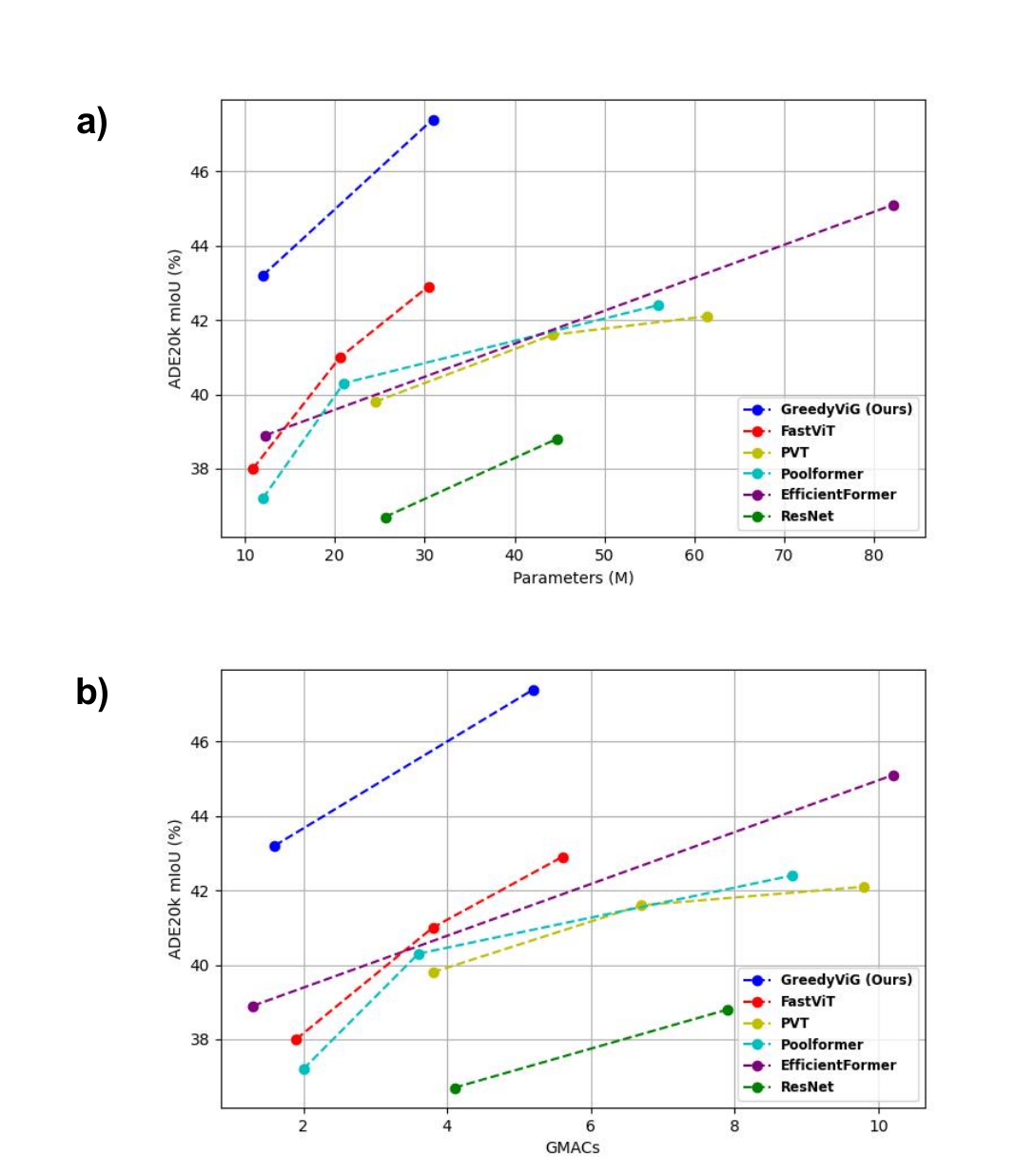}
\caption{\textbf{Comparison of model size and performance (\textit{mIoU} on ADE20K)}. GreedyViG achieves the highest performance on all model sizes compared to other state-of-the-art models. a) shows performance compared to parameters and b) shows performance compared to GMACs.} 
\label{fig:pareto_semantic}
\end{figure}

We further validate the performance of GreedyViG on semantic segmentation using the scene parsing dataset, ADE20k \cite{ADE20K}. The dataset contains 20K training images and 2K validation images with 150 semantic categories. Following the methodologies of \cite{MetaFormer, EfficientFormer, EfficientFormerv2, FastViT}, we build GreedyViG with Semantic FPN \cite{kirillov2019panoptic} as the segmentation decoder. The backbone is initialized with pretrained weights on ImageNet-1K and the model is trained for 40K iterations on 8 NVIDIA RTX 6000 Ada generation GPUs. We follow the process of existing works in segmentation, using the AdamW optimizer, set the learning rate as 2 $\times$ 10$^{-4}$ with a poly decay by the power of 0.9, and set the training resolution to 512 $\times$ 512 \cite{EfficientFormer, EfficientFormerv2}. 

As shown in Table \ref{Object_Detection_Segmentation_Results}, GreedyViG-S outperforms PoolFormer-S12 \cite{MetaFormer}, FastViT-SA12 \cite{FastViT}, and EfficientFormer-L1 \cite{EfficientFormer} by 6.0, 5.2, and 4.3 $mIoU$, respectively. Additionally, GreedyViG-B outperforms PoolFormer-S24 \cite{MetaFormer}, FastViT-SA36 \cite{FastViT}, and EfficientFormer-L3 \cite{EfficientFormer} by 7.1, 4.5, and 3.9 $mIoU$, respectively. Figure \ref{fig:pareto_semantic} shows GreedyViG significantly outperforms FastViT \cite{FastViT}, PVT \cite{wang2021pyramid}, Poolformer \cite{MetaFormer}, EfficientFormer \cite{EfficientFormer}, and ResNet \cite{Resnet} models with a similar number of parameters and GMACs.


\section{Ablation Studies}
\label{Sec:Ablations}

The ablation studies are conducted on ImageNet-1K \cite{imagenet1k}. Table \ref{tab:ablation2} reports the ablation study of GreedyViG-B on varying distances of considered graph connections ($K$) in the DAGC algorithm and how conditional positional encoding affects performance.

\begin{table}[ht]
\def\arraystretch{1.2}
\caption{\textbf{Ablation study for GreedyViG-B on ImageNet-1K benchmark} for varying distances between considered node connections (K) and the addition of conditional positional encoding.}
\centering
\begin{tabular}[t]{|c|c|c|c|}
\hline
\textbf{K} & \textbf{Params (M)} & \textbf{CPE} & \textbf{Top-1 (\%)}  \\ \hline
K = 16, 8, 4, 2   & 30.9    & Yes             & 83.5 \\
K = 9, 6, 3, 1   & 30.9     & Yes            & 83.7 \\
K = 8, 4, 2, 1     & 30.7   & No           & 83.7 \\
\rowcolor {Gray}
\textbf{K = 8, 4, 2, 1}   & \textbf{30.9}  &    \textbf{Yes}          &  \textbf{83.9}  \\ \hlineB{5}
\end{tabular}
\label{tab:ablation2}
\end{table}

\textbf {Distance between considered nodes for graph construction (\textit{K}).} The distance considered between possible node connections for graph construction can create a sparser graph, but can lead to decreased accuracy as the graph becomes too sparse and does not contain enough connections. We can see this in Table \ref{tab:ablation2}, which shows that for $K$ = 16, 8, 4, 2 in stages 1, 2, 3, and 4 that the top-1 accuracy is 0.4\% lower than for when $K$ = 8, 4, 2, 1. We also find that using $K$ = 9, 6, 3, 1 leads to a 0.2\% decrease in top-1 accuracy compared to $K$ = 8, 4, 2, 1 used in GreedyViG.

\textbf{Conditional Positional Encoding (CPE).} The encoding of positions within GreedyViG also boosts performance for relatively few parameters. When removing CPE from GreedyViG-B, we see a drop in accuracy of 0.2\% with only a 0.2 M decrease in parameters, showing that CPE is beneficial in the GreedyViG architecture.

Further ablation studies on the effects of removing graph convolutions at higher resolution stages, static versus dynamic graph construction, and how graph construction impacts latency are included in the supplementary material.


\section{Conclusion}
\label{Sec:Conclusion}

In this work, we have proposed a new method for designing efficient vision GNNs, Dynamic Axial Graph Construction (DAGC). DAGC is more efficient compared to KNN-based ViGs and more representative compared to SVGA. This is because DAGC uses an axial graph construction method to limit graph connections, and it does not have a fixed number of graph connections allowing for a variable number of connections based on the input image. Compared to past axial graph construction methods, DAGC limits the graph connections made within an image to only the significant connections thereby constructing a more representative graph. Additionally, we have proposed a novel CNN-GNN architecture, GreedyViG, which uses DAGC. GreedyViG outperforms existing ViG, CNN, and ViT models on multiple representative vision tasks, namely image classification, object detection, instance segmentation, and semantic segmentation. GreedyViG shows that ViG-based models can be legitimate competitors to ViT-based models through DAGC and by performing local and global processing at each resolution through a hybrid CNN-GNN architecture.

\section{Acknowledgements}
\label{Sec:Acknowledgement}

This work is supported in part by the NSF grant CNS 2007284, and in part by the iMAGiNE Consortium (\url{https://imagine.utexas.edu/}).

\noindent 

{
    \small
    \bibliographystyle{ieeenat_fullname}
    \bibliography{main}
}

\input{sec/X_suppl}

\end{document}

%% file: sec/X_suppl.tex
\clearpage
\setcounter{page}{1}
\maketitlesupplementary
\appendix

\section{Further Ablation Studies}
\label{Sec:Suppl_Ablations}

The ablation studies are conducted on ImageNet-1K \cite{imagenet1k}. Table \ref{tab:ablation_stages} reports the ablation study of GreedyViG-B (GViG-B) on the effects of graph convolutions at higher resolution stages and Table \ref{tab:ablation_construction} reports the effects of static versus dynamic graph construction.

\textbf{Graph convolutions at higher resolution stages.} In Table \ref{tab:ablation_stages} we can see that adding graph convolutions at higher resolution stages improves top-1 accuracy with a relatively small increase in parameters. By 1-stage, 2-stage, 3-stage, and 4-stage we mean that the DAGC blocks (graph convolution block) will be used in stage 4, stages 3 and 4, stages 2, 3, and 4, or in all stages as shown in Figure \ref{fig:GreedyViG_Architecture}. GreedyViG-B increases in top-1 accuracy as we move from 1-stage to 4-stage increasing from 83.1\% at the 1-stage configuration to 83.5\% at the 2-stage configuration. Moving from the 2-stage configuration to the 3-stage configuration we see a 0.2\% increase in accuracy reaching 83.7\%. Finally, moving from 3-stage to 4-stage we see a 0.2\% increase in accuracy reaching 83.9\%  top-1 accuracy at the 4-stage configuration. Comparing the 1-stage and 4-stage configurations we see a 0.8\% gain in top-1 accuracy with only an increase of 4.4 M parameters, showing the benefits of graph convolutions at higher resolution stages.

\begin{table}[H]
\footnotesize
\def\arraystretch{1.2}
\caption{\textbf{Ablation study for graph convolutions at higher resolution stages on ImageNet-1K benchmark.} 1-S, 2-S, 3-S, and 4-S indicate that graph convolutions were used in 1-stage, 2-stages, 3-stages, or all 4-stages. A check mark indicates this component was used in the experiment. A (-) indicates this component was not used.}

\centering
\begin{tabular}[t]{|c|c|c|c|c|c|c|c|c|c}
\hline
{\textbf{Model}} & {\textbf{Params (M)}} & {\textbf{1-S}} &{\textbf{2-S}} & {\textbf{3-S}} & {\textbf{4-S}} &{\textbf{Top-1 (\%)}} \\ \hline

GViG-B                     & 26.5  & \checkmark & - & - & - & 83.1 \\
GViG-B                     & 29.7  & - & \checkmark & - & - & 83.5 \\
GViG-B                    & 30.7   & - & - & \checkmark & - &  83.7 \\
\rowcolor {Gray}
\textbf{GViG-B}                   & \textbf{30.9} & - & - & - & \checkmark & \textbf{83.9}   \\ \hline
\end{tabular}
\label{tab:ablation_stages}
\end{table}

\textbf {Static versus dynamic graph construction.} Compared to the static graph construction method (SVGA) proposed in \cite{MobileViG}, DAGC connects only the similar connections based on Euclidean distance resulting in improved performance. In Table \ref{tab:ablation_construction} we can see the direct benefit of using DAGC compared to SVGA as it adds no parameters and increases the top-1 accuracy of GreedyViG-B with 4-stages by 0.4\% from 83.5\% to 83.9\%. We can also see the benefit of DAGC and our overall GreedyViG architecture compared to the MobileViG architecture, which uses SVGA, through comparing MobileViG-B (MViG-B) and a 1-stage configuration of GreedyViG-B. The 1-stage configuration of GreedyViG-B shows a 0.5\% improvement in top-1 accuracy from 82.6\% to 83.1\% while reducing parameters by 0.2 M, showing the benefits of dynamic graph construction.

\begin{table}[H]
\scriptsize
\def\arraystretch{1.2}
\caption{\textbf{Ablation study for static versus dynamic graph construction on ImageNet-1K benchmark.} 1-S indicates that graph convolutions were only used in Stage 4, while 4-S indicates that graph convolutions were used in stages 1, 2, 3, and 4. A check mark indicates this component was used in the experiment. A (-) indicates this component was not used.}

\centering
\begin{tabular}[t]{|c|c|c|c|c|c|c|c|c|c}
\hline
{\textbf{Model}} & {\textbf{Params}} & {\textbf{SVGA}} & {\textbf{DAGC}} & {\textbf{1-S}} & {\textbf{4-S}} & {\textbf{Top-1 (\%)}} \\ \hline

MViG-B \cite{MobileViG}              & 26.7 M & \checkmark & - & \checkmark & - & 82.6 \\
GViG-B                     & 26.5 M & - & \checkmark & \checkmark & - & 83.1 \\
GViG-B                    & 30.9 M & \checkmark  & -  & - & \checkmark &  83.5 \\
\rowcolor {Gray}
\textbf{GViG-B}                   & \textbf{30.9 M} & -    & \checkmark    & - & \checkmark & \textbf{83.9}   \\ \hline
\end{tabular}
\label{tab:ablation_construction}
\end{table}

\section{Network Configurations}
\label{Sec:Suppl_configuration}

The detailed network architectures for GreedyViG-S, M, and B are provided in Table \ref{table_of_arch}. We report the configuration of the stem, stages, and classification head. In each stage the number of MBConv and DAGC blocks repeated as well as their channel dimensions is reported. For GreedyViG-B, stage 4 has 3 repeated MBConv and DAGC blocks instead of 4 in order to have comparable parameters to other competing architectures.

\begin{table}[H]
\scriptsize
\caption{\textbf{Architecture details of GreedyViG} showing configuration of the stem, stages, and classification head. $C$ represents the channel dimensions.}
\centering
\setlength{\tabcolsep}{4pt}
\begin{tabular}{|c|c|c|c|c|c|}
\hline
Stage                             & GreedyViG-S & GreedyViG-M & GreedyViG-B \\ \hline \rule{0pt}{4ex}
Stem                                    & Conv $\times$2             & Conv $\times$2            & Conv $\times$2           \\[8pt] \hline \rule{0pt}{4ex}
Stage 1           & $ \begin{array}{ccc} MBConv \times2 \\ DAGC \times2 \\ C = 48 \end{array} $             & $ \begin{array}{ccc} MBConv \times3 \\ DAGC \times3 \\ C = 56 \end{array} $          
& $ \begin{array}{ccc} MBConv \times4 \\ DAGC \times4 \\ C = 64 \end{array} $ 
 \\[8pt] \hline \rule{0pt}{4ex}           
Stage 2           & $ \begin{array}{ccc} MBConv \times2 \\ DAGC \times2 \\ C = 96 \end{array} $             & $ \begin{array}{ccc} MBConv \times3 \\ DAGC \times3 \\ C = 112 \end{array} $          
& $ \begin{array}{ccc} MBConv \times4 \\ DAGC \times4 \\ C = 128 \end{array} $
\\[8pt] \hline \rule{0pt}{4ex}
Stage 3              & $ \begin{array}{ccc} MBConv \times6 \\ DAGC \times2 \\ C = 192 \end{array} $             & $ \begin{array}{ccc} MBConv \times9 \\ DAGC \times3 \\ C = 224 \end{array} $          
& $ \begin{array}{ccc} MBConv \times12 \\ DAGC \times4 \\ C = 256 \end{array} $
\\[8pt] \hline \rule{0pt}{4ex}
Stage 4                   & $ \begin{array}{ccc} MBConv \times2 \\ DAGC \times2 \\ C = 384 \end{array} $             & $ \begin{array}{ccc} MBConv \times3 \\ DAGC \times3 \\ C = 448 \end{array} $          
& $ \begin{array}{ccc} MBConv \times3 \\ DAGC \times3 \\ C = 512 \end{array} $           \\[8pt] \hline 
Head                                         & Pooling \& MLP            & Pooling \& MLP            & Pooling \& MLP            \\ \hline
\end{tabular}
\label{table_of_arch}
\end{table}

\section{Computational Complexity of Graph Construction}
\label{Sec:Computational_complexity}

The computational complexity for KNN, DAGC, and SVGA for a single node in the image (in terms of comparisons from that node) is given below. $W$ and $H$ are the width and height of the image, $K$ is the number of nearest neighbors, and $N$ is the number of hops selected in SVGA and DAGC.

\begin{enumerate}
    \item \textbf{KNN}: O($W \times H \times K$). For each node, KNN finds the $K$ nearest by comparing every node to the current node.

    \item \textbf{DAGC}: O($\frac{W + H}{N}$). For each node, DAGC only needs to compare nodes that are every $N$ hops away, thus decreasing the number of comparisons. Also, since DAGC computes the $\mu$ and $\sigma$ beforehand, it makes connections in the first search through of the image rather than needing to compare again for $K$ connections.

    \item \textbf{SVGA}: O(1). Connects each node along the axes.
\end{enumerate}

DAGC is more computationally expensive than SVGA, but more representative. KNN may be more representative than DAGC, but can cause oversmoothing and is more computationally expensive. The measured time taken for graph construction is 0.06 ms in DAGC, 0.38 ms in KNN, and 0.04 ms in SVGA when measured on an Nvidia RTX A6000; this shows DAGC is slower than SVGA and faster than KNN in graph construction time. This can also be seen through our latency results in Table \ref{tab:latency_graph_construction}. GreedyViG-S is faster and more accurate than PViG-Ti, but is slower and more accurate than a smaller MobileViG-S model. GreedyViG-S is slower compared to MobileViG-S because DAGC is slower than SVGA, GreedyViG has more parameters, and because GreedyViG contains more global processing stages that perform graph convolution (DAGC blocks) as compared to MobileViG which only does graph convolution at its lowest resolution stage after multiple downsample layers.

\begin{table}[ht]
\footnotesize
\def\arraystretch{1.2}
\caption{\textbf{Graph construction impact on accuracy and latency.} We show GreedyViG-S with KNN and DAGC to compare with PViG-Ti with KNN and DAGC. We also show MobileViG-S with SVGA to show it is less accurate, but faster than GreedyViG-S.}
\centering
\begin{tabular}[t]{|c|c|c|c|}
\hline
\textbf{Model} & \textbf{Params} & \textbf{Latency} & \textbf{Acc (\%)}  \\ \hline
MobileViG-S \cite{MobileViG} w/ SVGA  & 7.2 M   & 27.1 ms           & 78.2 \\
PViG-Ti \cite{Vision_GNN} w/ KNN  & 10.7 M        & 79.4 ms     & 78.2 \\
PViG-Ti \cite{Vision_GNN} w/ DAGC   & 10.7 M      & 63.3 ms   & 79.1  \\
GreedyViG-S (Ours) w/ KNN  & 12.0 M    & 73.6 ms  & 80.2 \\
\textbf{GreedyViG-S (Ours) w/ DAGC}  & \textbf{12.0 M}   & \textbf{53.4 ms}  & \textbf{81.1} \\ \hlineB{5}
\end{tabular}
\label{tab:latency_graph_construction}
\end{table}


The graph construction and architecture of GreedyViG both contribute to the performance of GreedyViG models. When using DAGC with the original ViG architecture and KNN with our GreedyViG architecture in Table \ref{tab:latency_graph_construction}, we can see that DAGC is faster and provides higher accuracy compared to KNN in these cases. GreedyViG-B with SVGA can also be seen Table \ref{tab:ablation_construction}, showing with the same configuration DAGC has 83.9\% accuracy compared to SVGA's 83.5\%.